\documentclass[final]{cvpr}

\usepackage{times}
\usepackage{epsfig}
\usepackage{graphicx}
\usepackage{amsmath}
\usepackage{amssymb}
\usepackage{algpseudocode}
\usepackage{floatrow}
\usepackage{subcaption}
\usepackage{algorithmicx}
\usepackage{algorithm}
\usepackage{siunitx}
\newcommand\sbullet[1][.5]{\mathbin{\vcenter{\hbox{\scalebox{#1}{$\bullet$}}}}}
\newcommand\Tau{\mathcal{T}}

\usepackage[pagebackref=true,breaklinks=true,letterpaper=true,colorlinks,bookmarks=false]{hyperref}

\usepackage[pagebackref=true,breaklinks=true,colorlinks,bookmarks=false]{hyperref}

\author{Jiangpeng He\\
{\tt\small he416@purdue.edu}
\and
Fengqing Zhu\\
{\tt\small zhu0@purdue.edu}
\and
{School of Electrical and Computer Engineering, Purdue University, West Lafayette, Indiana USA}
}

\begin{document}

\title{Online Continual Learning For Visual Food Classification}

\author{Jiangpeng He\\
{\tt\small he416@purdue.edu}
\and

Fengqing Zhu\\
{\tt\small zhu0@purdue.edu}
\and
{School of Electrical and Computer Engineering, Purdue University, West Lafayette, Indiana USA}
}

\maketitle

\begin{abstract}
Food image classification is challenging for real-world applications since existing methods require static datasets for training and are not capable of learning from sequentially available new food images. Online continual learning aims to learn new classes from data stream by using each new data only once without forgetting the previously learned knowledge. However, none of the existing works target food image analysis, which is more difficult to learn incrementally due to its high intra-class variation with the unbalanced and unpredictable characteristics of future food class distribution. In this paper, we address these issues by introducing (1) a novel clustering based exemplar selection algorithm to store the most representative data belonging to each learned food for knowledge replay, and (2) an effective online learning regime using balanced training batch along with the knowledge distillation on augmented exemplars to maintain the model performance on all learned classes. Our method is evaluated on a challenging large scale food image database, Food-1K\footnote{https://www.kaggle.com/c/largefinefoodai-iccv-recognition/data}, by varying the number of newly added food classes. Our results show significant improvements compared with existing state-of-the-art online continual learning methods, showing great potential to achieve lifelong learning for food image classification in real world.


\end{abstract}


\section{Introduction}
\label{introduction}

\begin{figure}[t]
\begin{center}
  \includegraphics[width=1.\linewidth]{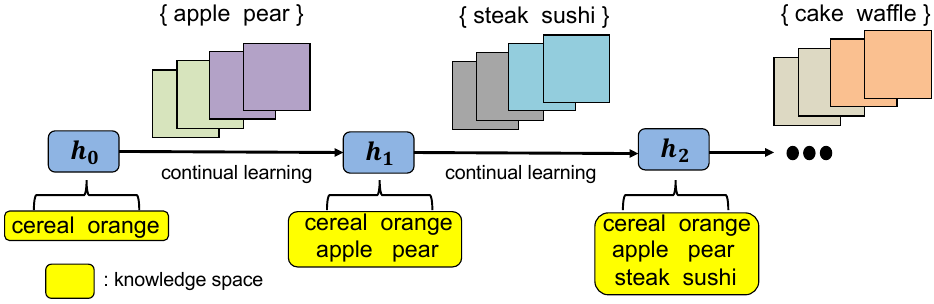}
  \vspace{-0.8cm}
  \caption{\textbf{Continual learning for food image classification.} The model $h$ learns new food class sequentially overtime without accessing to already learned class data for each continual learning step. The updated model can classify all food classes seen so far.}
  \label{fig:intro}
\end{center}
\end{figure}

Food classification serves as the first and most crucial step for image-based dietary assessment~\cite{boushey2017new}, which aims to provide valuable insights for prevention of many chronic diseases. As shown in Figure~\ref{fig:intro}, ideally food classification system should be able to update using each new recorded food image continually without forgetting the food class that has been already learned before. Achieving this goal would bring significant advantage for deploying such a system for automated dietary assessment and monitoring.

From the perspective of visual food classification, although recent works~\cite{wu2016learning,mao2020visual,Food2K,min2020isia-500} have been proposed using advanced deep learning based approaches to increase model performance, they use only static datasets for training and are not capable of handling sequentially available new food classes. Therefore, the classification accuracy could drop dramatically due to the unavailability of old data, which is also known as catastrophic forgetting~\cite{CF}. Although retraining from scratch is a viable option, it is impractical to do whenever a new food is observed, which is time consuming and require high computation and memory resource especially for large scale food image datasets. For example, a model already learned $1,000$ food classes need to retrain from scratch for only $1$ new observed food.

From the perspective of continual learning, an increasing number of approaches~\cite{ILIO, ER_1, ER_2, prabhu2020gdumb_online} have been proposed to address catastrophic forgetting and to learn new knowledge incrementally in online scenario. Compared to offline scenario where data can be used multiple epochs for training, online scenario is more challenging where each new data is observed only once by the model, but is more practical for real-life application such as food image classification system. Representative techniques to mitigate forgetting include (1) storing a small number of learned data as exemplars for replay~\cite{ICARL}, and (2) applying knowledge distillation~\cite{KD} using a teacher model to maintain the learned performance. However, continual learning for food image classification is still lacking and there are two major obstacles which make the above mentioned techniques less effective for food images. (i) Food images exhibit higher intra-class variation~\cite{mao2020visual} compared with commonly seen objects in real life, which is due to different culinary culture and cooking style. Most existing continual learning methods~\cite{ICARL,EEIL,BIC,mainatining,ILIO,rebalancing} apply herding algorithm~\cite{HERDING} to select exemplars for each learned class based on class mean only, which is difficult to cover the diversity for food types within the same class. Therefore, catastrophic forgetting could become worse if stored exemplars are not good representations of learned classes. (ii) The distribution of future food classes is usually unpredictable and imbalanced due to the variance of consumption frequencies~\cite{lin2021most} among different food categories. Nevertheless, most online approaches only study continual learning on balanced datasets containing the same number of data per class such as CIFAR~\cite{CIFAR} and MNIST~\cite{MNIST} without considering the class-imbalance problem that is common for food images. In addition, as indicated in~\cite{dualmemory}, the knowledge distillation term becomes less effective if teacher model is not trained on balanced data. 

In this work, we address the challenging problem of food image classification for online continual learning by first introducing a novel exemplar selection algorithm, which clusters data for each class based on visual similarity and then selects the most representative exemplars from each generated cluster based on cluster mean. We apply Power Iteration Clustering~\cite{PIC}, which does not require the number of cluster beforehand. Therefore, our algorithm can adapt to different food categories, \textit{i.e.}, food with higher variation will generate more clusters and vice versa. In addition, we propose an effective online learning regime by using balanced training batch for old and new class data and apply knowledge distillation loss between original and augmented exemplars to better maintain the model performance. Our method is evaluated on a large scale real world food database, Food-1K~\cite{Food2K}, and outperforms state-of-the-arts including ICARL~\cite{ICARL}, ER~\cite{ER_1, ER_2}, ILIO~\cite{ILIO} and GDUMB~\cite{prabhu2020gdumb_online}, which are all implemented in online scenario and use exemplars for replay during continual learning.

The main contributions are summarized as follows.
\begin{itemize}
  \item To the best of our knowledge, we are the first to study online continual learning for food image classification. We propose a novel clustering based exemplar selection algorithm and a new online training regime to address catastrophic forgetting.
  
  \item We conduct extensive experiments on a challenging class-imbalanced food image database to show the effectiveness for each component of our proposed method. We show that our method significantly outperforms existing approaches, especially for larger incremental step size.
  
\end{itemize}


\section{Related Work}
\label{related work}
\subsection{Food Classification}
Food classification refers to the task of labeling image with food category, which assumes each input image contains only one single food item. Earlier work~\cite{hand_crafted2} use fusion of features including SIFT~\cite{SIFT}, Gabor, and color histograms for classification. Later, the modern deep learning models have been widely applied as backbone network to extract more class-discriminative features as in~\cite{Kagaya, Tanno2016, wide, Food2K, he2020multitask, he2021end, min2020isia-500, shao2021towards}, which significantly improves the performance. Recent works~\cite{wu2016learning, mao2020visual} leveraging hierarchy structure based on visual information are able to achieve further improvements. However, all these methods use static food image datasets for training and none of them is capable of learning from sequentially available data, making it difficult to apply in real life applications as new foods are observed over time.

\begin{figure*}[htbp]
\begin{center}
  \includegraphics[width=.95\linewidth]{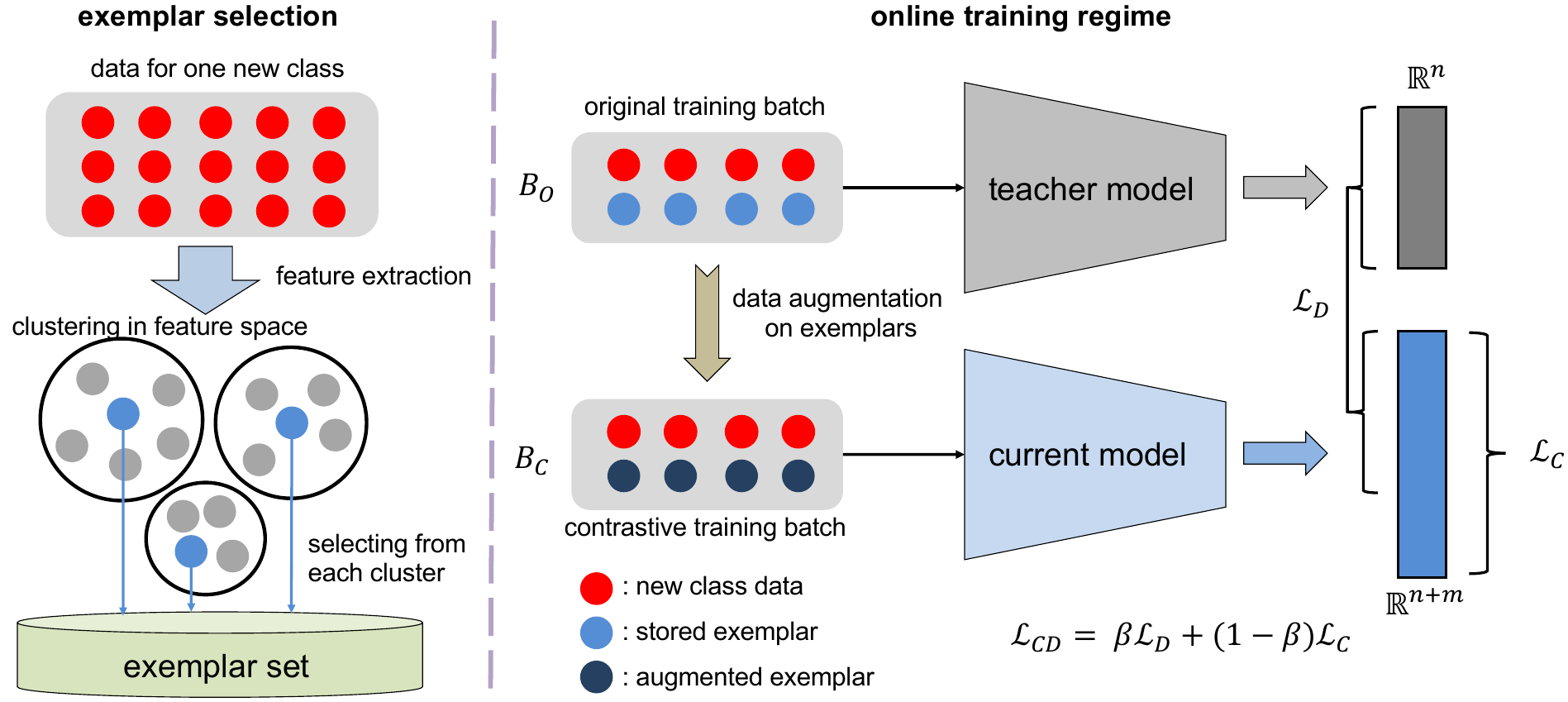}
  \caption{\textbf{Overview of proposed method.} The left side shows our exemplar selection algorithm, which selects the most representative data from center of each cluster generated based on visual similarity in feature space. Right part shows our online learning regime where each new class data is paired with one randomly selected exemplar to produce the original balanced training batch $B_O$. We perform data augmentation on selected exemplars to generate contrastive training batch $B_C$ and the distillation loss $\mathcal{L}_D$ is applied between the output of the teacher model using $B_O$ and the output of the current model using $B_C$. $n$ and $m$ denote the number of already learned classes and new added classes, respectively. $\beta$ is a hyper-parameter to combine $\mathcal{L}_D$ with cross-entropy loss $\mathcal{L}_C$.  (Best viewed in color)}
  \label{fig:overall}
\end{center}
\end{figure*}

\subsection{Continual Learning}
The major challenge for continual learning is called catastrophic forgetting~\cite{CF}, where the model quickly forgets already learned knowledge due to the unavailability of old data. Below, we review and summarize existing knowledge-preserving techniques that are most relevant to our proposed method. 

\textit{Replay-based} methods store a small number of representative data from each learned class as exemplars to perform knowledge rehearsal during the continual learning. Herding dynamic algorithm~\cite{HERDING} is first applied in ICARL~\cite{ICARL} to select exemplars that are closer to the class mean. It has gradually became a common exemplar selection strategy that is being used in most existing methods~\cite{ICARL,EEIL,BIC,mainatining,ILIO,rebalancing}, where ICARL adopts a nearest class mean classifier~\cite{NCM_classifier} while others use softmax classifier for classification. In addition, reservoir sampling~\cite{vitter1985random} along with random retrieval is applied in Experience Replay (ER) based methods~\cite{ER_1,ER_2}, which ensures each incoming data point has the same probability to be selected as exemplar in the memory buffer. A greedy balancing sampler with random selection is recently used in GDUMB~\cite{prabhu2020gdumb_online} to store as much data as memory allowed, which also achieves competitive performance. 

\textit{Regularization-based} methods restrict the impact of learning new tasks on the parameters that are important for learned tasks. Knowledge distillation~\cite{KD} is a popular representative technique, which makes the model mimic the output distribution for learned classes from a teacher model to mitigate forgetting during continual learning~\cite{LWF,ICARL,EEIL,BIC,rebalancing,inthewild,he2021unsupervised}. For most recent work, He \textit{et al}. proposed ILIO~\cite{ILIO}, which applies an accommodation ratio to generate a stronger constraint for knowledge distillation loss to achieve improved performance. 

However, among these methods, only a few~\cite{ICARL,ILIO,prabhu2020gdumb_online,ER_1,ER_2} are feasible in online scenario to use each data only once for training. In addition, none of the existing methods focus on food images and as introduced in Section~\ref{introduction}, the high intra-class variance and imbalanced data distribution make both exemplar and distillation based techniques less effective to address catastrophic forgetting. Therefore, we propose a novel exemplar selection algorithm to select exemplars from each generated cluster based on visual similarity to adapt to the variability of different food categories. Besides, we propose an effective online learning regime using balanced training batch and apply distillation on augmented exemplars to better maintain performance on learned classes, which is described in details in Section~\ref{method:our}.

\section{Problem Statement For Online Continual Learning}
\label{Problem Statement}

Continual learning has been studied under different scenarios. In general, it can be divided into (1) task-incremental (2) class-incremental and (3) domain-incremental as discussed in~\cite{hsu2018re}. Methods for task-incremental problem use a multi-head classifier~\cite{abati2020conditional_taskaware} for each independent task while task index is not available in class-incremental problem, which applies a single-head classifier~\cite{SIT} on all learned classes. Domain-incremental aims to learn the label shift instead of new classes. In addition, depending on whether each data is allowed to use more than once to update model, it can be categorized into (1) online learning that use each data once and (2) offline learning with no epoch restriction. In this work, we focus on online continual learning under class-incremental setting, which is more related to real life applications. The objective is to learn new class from data stream using each data once and to classify all classes seen so far during inference. 

Specifically, the online class-incremental learning problem $\Tau$ can be formulated as learning a sequence of $N$ tasks $\{\Tau^1,...,\Tau^N\}$ corresponds to $N$ incremental learning steps with model updating from $h^0$ to $h^N$, where the initial model $h^0$ is assumed to be trained on $\Tau^0$ before continual learning begins and $h^N$ should be able to perform classification on test data belonging to $\{\Tau^0,\Tau^1...,\Tau^N\}$. Each task $\Tau^i \in \Tau$ for $i = \{0,1,...N\}$ contains fixed $M$ non-overlapped new classes, which is defined as incremental step size. Let $\{D^0, D^1, ..., D^N\}$ denotes training data corresponds to $N$ incremental steps plus the initial step $D^0$, where $D^i = \{(\textbf{x}_1^i,y_1^i)...(\textbf{x}_{n_i}^i,y_{n_i}^i)\}$, $\textbf{x}$ and $y$ represent the data and the label respectively, and $n_i$ refers to the number of total training data in $D^i$. In online scenario, the new class data for each incremental learning step becomes available sequentially and one does not need to wait until all data has arrived to update the model as in offline case. The online learner observes each data $(\textbf{x}^i,y^i)\in D^i$ only once for incremental step $i$.

\section{Our Method}
\label{method:our}
An overview of our proposed method is illustrated in Figure~\ref{fig:overall}, including a novel exemplar selection method and an effective online training regime. Specifically, instead of selecting exemplars based on class mean as in herding~\cite{HERDING}, we first generate clusters based on similarity and then select exemplars from each cluster using the corresponding cluster mean. During the continual learning phase, each new class data from data stream is paired with one randomly selected exemplar from exemplar set to produce balanced training batch $B_o$ that contains the same number of original new and old class samples. Then we apply data augmentation on selected exemplars in $B_o$ to generate a contrastive training batch $B_c$ and the knowledge distillation term is applied between the teacher output of $B_o$ and the current model output of $B_c$ to maintain the already learned knowledge. Details of each component is described in the remaining section. 

\subsection{Exemplar Selection From Clusters}
\label{method:exp selection}
The main challenge of existing exemplar selection methods is that they cannot adapt to the intra-class variation especially for food images due to its high variability. For example, the images in apple category may contain many types such as green apple, red apple, sliced apple, diced apple, whole apple and etc. Therefore, selecting from class mean as in Herding~\cite{HERDING} will not work well when there exists more than one main types within that food class. Our proposed method addresses this problem by first clustering the data for each class based on visual similarity and then select exemplars from each generated cluster. We consider Power Iteration Clustering (PIC)~\cite{PIC} as our clustering approach, which is a graph based method and shown to be effective even in large scale database~\cite{douze2017evaluation}. But other clustering methods are also feasible such as K-means~\cite{K-MEANS}. One advantage of PIC is the number of generated clusters are not set beforehand, so there is more clusters if one class contains more main types and vice versa. 

Given $n_c$ images $\{(\textbf{x}_1,y),... (\textbf{x}_{n_c},y)\}$ for one new class $c$, we first generate nearest neighbor graph by connecting to their $10$ neighbor data points in the Euclidean space using extracted feature embeddings. Let $f(\textbf{x}_i)$ denotes the extracted feature for the $i$-th image, we apply the sparse graph matrix $G = \mathbb{R}^{n_c\times n_c}$ with zeros on the diagonal and the remaining elements of $G$ are defined by $$e_{i,j} = exp^-\frac{||f(\textbf{x}_i) - f(\textbf{x}_j)||^2}{\sigma^2}$$ where $\sigma$ denotes the bandwidth parameter and we empirically use $\sigma = 0.5$ in this work. Then, we initialize a starting vector $s^{n_c\times1} = [\frac{1}{n_c}, ..., \frac{1}{n_c}]^T$ and iteratively update it using Equation~\ref{eq:pic}
\begin{equation} \label{eq:pic}
\begin{aligned}
s = L_1(\alpha(G + G^t)s + (1-\alpha)s)
\end{aligned}
\end{equation}
where $\alpha = 0.001$ refers to a regularization parameter and $L_1(\sbullet[.5])$ denotes the L-1 normalization step. The generated clusters are given by the connected components of a directed unweighted subgraph of $G$ denoted as $\tilde{G}$. We set $\tilde{G}_{i,j} = 1$ if $j = \textit{argmax}_je_{i,j}(s_j - s_i)$ where $s_i$ refers to the $i$-th element of the vector. Note that there is no edge starts from $i$ if $\{\forall j\neq i, s_j \leq s_i\}$, \textit{i.e. }$s_i$ is a local maximum.

The general process to select exemplars for a new class $c$ after incremental step $k$ is illustrated in Algorithm~\ref{alg:exemplar selection}, where we select the same number of exemplars $q_e$ from each cluster generated using PIC based on cluster mean. Note that for the special situation when a cluster $C_i$ contains very few data with $|C_i| < q_e$,  we store all data from that small cluster at first and then evenly select from the remaining clusters.

\begin{algorithm}[t]
\caption{Selecting exemplars for a new class $c$}
\begin{flushleft}
    \hspace*{0.02in} {\bf Input:}
    New class data: $\{(\textbf{x}_1,y), ... (\textbf{x}_{n_c},y)\} \in \Tau^k $\\
    \hspace*{0.02in} {\bf Require:} 
    Clustering algorithm : $\Theta$\\
    \hspace*{0.02in} {\bf Require:} 
    Number of exemplars per class : q \\
    \hspace*{0.02in} {\bf Output:} 
    Exemplar set for new class : $E_c$\\
\end{flushleft}
\vspace{-0.35cm}
\begin{algorithmic}[1]
\State $E_c \leftarrow \emptyset$ \Comment{\small{initialization of exemplar set for new class $c$}} 
\State $f \leftarrow h^k$ \Comment{\small{use current model as feature extractor}} 
\State $C_1, ...C_n \leftarrow  \Theta(f(\textbf{x}_1),...f(\textbf{x}_{n_c}))$  \Comment{\small{generated clusters}} 
\State $q_e \leftarrow floor(\frac{q}{n})$ \Comment{\small{number of exemplar for each cluster}} 
\For{i = 1, 2, ... n}
\State $\mu_i = \frac{1}{|C_i|}\sum_{\textbf{x}\in C_i} f(\textbf{x})$
\Comment{\small{cluster mean}} 

\For{j = 1, 2, ... $q_e$}
\State $v_j \leftarrow \textit{argmin}_{\textbf{x}\in C_i} ||\mu_i - f(\textbf{x})||_2$
\State $E_c \leftarrow E_c \cup v_j $
\State $C_i \leftarrow C_i - v_j$ \Comment{\small{remove stored exemplar from cluster}} 
\EndFor
\EndFor
\end{algorithmic}
\label{alg:exemplar selection}
\end{algorithm}

\vspace{-0.1cm}
\subsection{Online Learning Regime}
\vspace{-0.1cm}
\label{method:training regime}
Since future food class distribution is usually unpredictable and imbalanced, it becomes more challenging to maintain the learned knowledge due to potential class-imbalanced problem. However, almost all existing online continual learning methods use balanced datasets such as MNIST~\cite{MNIST} and CIFAR~\cite{CIFAR} which contain the same number of training data for each class. In addition, the knowledge distillation term also becomes less effective when the teacher model is not trained on balanced data~\cite{dualmemory}. Therefore, we propose a more effective online learning regime, which consists of two main parts: using balanced training batch and applying knowledge distillation on augmented exemplars. 

Suppose the model is already trained on $n$ classes and the data stream $ \{(\textbf{x}_1^k,y_1^k)...\} \in D^k $ for incremental step $k$ contains $m$ newly added classes where $y^k\in\{n+1, n+2,..., n+m\}$. We pair each new class data $(\textbf{x}_i^k,y_i^k)$ with a randomly selected exemplar $(\textbf{v}_j, y_j) \in E^{k-1}$ where $E^{k-1}$ denotes exemplar set containing stored exemplars for classes $\{1,2,...,n\}$ belonging to $\{\Tau^0,...,\Tau^{k-1}\}$. Therefore, each training batch $B$ contains exactly $\frac{b}{2}$ new class data and $\frac{b}{2}$ augmented old class exemplars given batch size $b = |B|$.

To make the distillation term more effective, instead of using the identical training batch for both current model and teacher model as done in existing approaches, we propose to apply data augmentation on selected exemplars in original training batch $B_o$ to generate its corresponding contrastive training batch $B_c$ where $B_c$ and $B_o$ are used as input to current model and teacher model, respectively.

The output logits of the current model is denoted as $p^{(n+m)}(B_c(\textbf{x})) = (o^{(1)},...,o^{(n)},o^{(n+1)}, ...o^{(n+m)})$, the teacher's output logits is $\hat{p}^{(n)}(B_o(\textbf{x})) = (\hat{o}^{(1)},...,\hat{o}^{(n)})$ where $B_c(\textbf{x})$ and $B_o(\textbf{x})$ denote the data in augmented and original training batch. The knowledge distillation loss~\cite{KD} is formulated as in Equation~\ref{eq:kd1}, where $\hat{p}_T^{(i)}$ and $p_T^{(i)}$ are the $i$-th distilled output logit as defined in Equation~\ref{eq:kd2}
\begin{equation} \label{eq:kd1}
\begin{aligned}
\mathcal{L}_{D}(B_c(\textbf{x}),B_o(\textbf{x})) = \sum_{i=1}^n-\hat{p}_T^{(i)}(B_o(\textbf{x}))log[p_T^{(i)}(B_c(\textbf{x}))]
\end{aligned}
\end{equation}
\begin{equation} \label{eq:kd2}
\begin{aligned}
\hat{p}_T^{(i)} =
\frac{\exp{(\hat{o}^{(i)}}/T)}{\sum_{j=1}^n\exp{(\hat{o}^{(j)}}/T)}\ , \ 
p_T^{(i)} = \frac{\exp{(o^{(i)}}/T)}{\sum_{j=1}^n\exp{(o^{(j)}}/T)}
\end{aligned}
\end{equation}
$T>1$ is the temperature scalar used to soften the distribution, which forces the network to learn more fine grained knowledge. The cross entropy loss to learn new classes can be expressed as in Equation~\ref{eq:cn}
\begin{equation} \label{eq:cn}
\begin{aligned}
\mathcal{L}_{C}(B_c(\textbf{x})) = \sum_{i=1}^{n+m}-\hat{y}^{(i)}log[p^{(i)}(B_c(\textbf{x}))]
\end{aligned}
\end{equation}
where $\hat{y}$ is the one-hot label for input data $x$. 
The overall cross-distillation loss function is formed as in Equation~\ref{eq:cd} by using a hyper-parameter $\beta$ to tune the influence between two components. 
\begin{equation} \label{eq:cd}
\begin{aligned}
\mathcal{L}_{CD}(B_c(\textbf{x})) = \beta \mathcal{L}_{D}(B_c(\textbf{x}),B_o(\textbf{x})) + (1-\beta) \mathcal{L}_{C}(B_c(\textbf{x}))
\end{aligned}
\end{equation} 

In this work, we set $T=2$ and $\beta = 0.5$. We also notice that using stronger random data augmentation techniques to generative contrastive training batch can achieve better performance to maintain the knowledge for learned classes. Therefore our data augmentation pipeline includes \textit{random flip}, \textit{random color distortions} and \textit{random Gaussian blur}.

\begin{table*}[t]
    \centering
    \scalebox{1.}{
    \begin{tabular}{|ccccccccccc|cc|}
        \hline
        Datasets & \multicolumn{10}{c}{\textbf{Food1K-100}} & \multicolumn{2}{|c|}{\textbf{Food1K}}\\
        \hline
        Step size& \multicolumn{2}{c}{1} & \multicolumn{2}{c}{2} & \multicolumn{2}{c}{5} & \multicolumn{2}{c}{10} & \multicolumn{2}{c}{20} & \multicolumn{2}{|c|}{100}\\
        \hline
        Accuracy & Avg & Last & Avg & Last & Avg & Last & Avg & Last & Avg & Last & Avg & Last\\
        \hline
        \hline
        Fine-tune & 0.043 & 0.009& 0.081& 0.029& 0.182 & 0.018 & 0.379 & 0.134 & 0.497 & 0.233 & 0.265 & 0.099\\
        Upper-bound & 0.805 & 0.759& 0789& 0.752& 0.807 & 0.743 & 0.827 & 0.749 & 0.813 & 0.744 & 0.788 & 0.805\\
        \hline
        ICARL~\cite{ICARL}  & 0.619 & 0.539 & 0.694 & 0.615 & 0.581 &0.502 & 0.729 & 0.603 & 0.769 & 0.660 & 0.573 & 0.474\\
        ER~\cite{ER_1,ER_2} & 0.645 & 0.586 & 0.612& 0.582 & 0.528 & 0.520 & 0.694 & 0.599 & 0.728 & 0.633 & 0.533 & 0.428\\
        GDUMB~\cite{prabhu2020gdumb_online} & 0.606 & 0.430 & 0.612& 0.441 & 0.573 & 0.507 & 0.591 & 0.456& 0.754 & 0.623 & 0.506 & 0.289 \\
        ILIO~\cite{ILIO} & \textbf{0.695} & \textbf{0.670}& 0.681 &\textbf{0.643} & 0.501 &0.452  & 0.703 & 0.633 & 0.708 & 0.596 & 0.515 & 0.428\\
        \hline
        Ours  & 0.692 & 0.661 & \textbf{0.702}&0.641 & \textbf{0.643} & \textbf{0.563 }&\textbf{0.762} & \textbf{0.669} & \textbf{0.786} & \textbf{0.699} & \textbf{0.612} & \textbf{0.504} \\
        \hline
    \end{tabular}
    }
    \vspace{-0.2cm}
    \caption{\textbf{Average accuracy and Last step accuracy} with step size 1, 2, 5, 10, 20 on Food1K-100 and step size 100 on Food-1K. Best results (except upper-bound) are marked in bold. }
    \label{tab:sota}
    
\end{table*}

\begin{figure*}[t!]
\begin{center}
  \includegraphics[width=1.\linewidth]{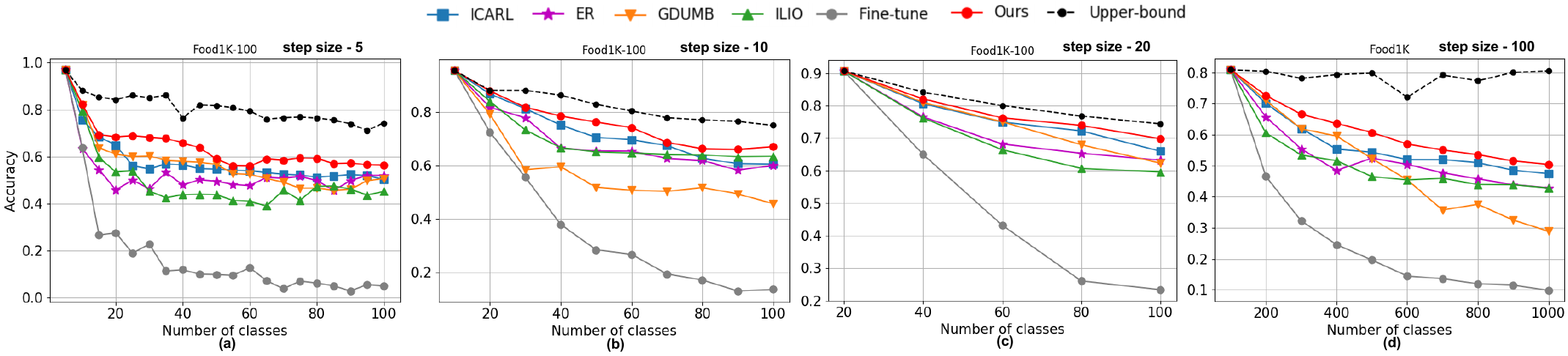}
  \vspace{-.9cm}
  \caption{\textbf{Accuracy for each incremental step} with step size (a) 5 (b) 10 (c) 20 on Food1K-100 and (d) step size 100 on Food-1K. (Best viewed in color)}
  \vspace{-0.3cm}
  \label{fig:sota}
\end{center}
\end{figure*}

\section{Experimental Results}
\label{experimental results}
In this section, we first compare our proposed online continual learning method with existing approaches including \textbf{ICARL}~\cite{ICARL}, \textbf{ER}~\cite{ER_1,ER_2}, \textbf{GDUMB}~\cite{prabhu2020gdumb_online} and \textbf{ILIO}~\cite{ILIO}, which all have already been discussed in Section~\ref{related work}. We also include \textbf{Fine-tune} and \textbf{Upper-bound} for comparison. \textbf{Fine-tune} use only new class data and apply cross-entropy loss for continual learning without considering the previous task performance, \textit{i.e.}, neither exemplar set nor distillation loss is used and it can be regarded as the lower-bound. \textbf{Upper-bound} trains a model using all the data seen so far for each incremental learning step using cross-entropy loss in online scenario. Results are discussed in Section~\ref{sec:compare with SOTA}. 

In the second part of this section, we conduct ablation study to show the effectiveness of each component of proposed method including exemplar selection algorithm and online training regime, which is illustrated in Section~\ref{sec: ablation study}.

\subsection{Datasets}
\label{sec:datasets}
In this work, we use \textbf{Food1K} to evaluate our method, which is a recently released challenging food dataset consisting of $1,000$ selected food classes from Food2K~\cite{Food2K}. The dataset is originally divided as $60\%$, $10\%$ and $30\%$ for training, validation and testing, respectively. Note that no class label is given in test set so we use images in validation set as testing data. In addition, we also construct a subset of Food1k using 100 randomly selected food classes denoted as \textbf{Food1K-100} for experiment. Specifically, for \textbf{Food1K-100}, we randomly arrange 100 classes into the splits of 1, 2, 5, 20 as step size (number of new class added for each step) and for \textbf{Food1K} we perform large scale continual learning using 100 new classes for each incremental step. 

\subsection{Implementation Details}
\label{sec:implementation details}
Our implementation is based on Pytorch~\cite{pytorch}. We use ResNet-18 as our backbone network by following the setting suggested in~\cite{RESNET} with input image size $224 \times 224$. We use stochastic gradient descent optimizer with fixed learning rate of $0.1$ and weight decay of $0.0001$. We store $q = 20$ exemplars per class in exemplar set as suggested in~\cite{ICARL} and the batch size is set as 32 (with 16 new class data paired with 16 randomly selected exemplars). For all experiments, each data (except stored exemplars) is used only once to update the model in online scenario. 

\textbf{Evaluation protocol:} after each incremental learning step, we evaluate the updated model on test data belonging to all classes seen so far and we use Top-1 accuracy for Food1K-100 and Top-5 accuracy for Food1K. Besides, we also report average accuracy (\textit{Avg}) and last step accuracy (\textit{Last}) for comparison where \textit{Avg} is calculated by averaging the accuracy for all incremental steps to show the overall performance for entire continual learning process and \textit{Last} accuracy shows the final performance on the entire dataset after the last step of continual learning. We repeat each experiment 5 times using different random seeds to arrange class and the average results are reported.

\subsection{Comparison With Existing Methods}
\label{sec:compare with SOTA}
Table~\ref{tab:sota} summarizes the average accuracy (\textit{Avg}) and last step accuracy (\textit{Last}) for all incremental step sizes. Overall, we notice that the online continual learning performance vary a lot for different step sizes. Given fixed total number of classes to learn, smaller step size will produce more incremental steps so catastrophic forgetting appears more frequently. On the other hand, for larger step size, although there will be less incremental steps, learning more classes for each step is also a challenging task especially in online scenario to use each data only once for training. Specifically, we observe severe catastrophic forgetting problem by using \textit{Fine-tune} where both \textit{Avg} and \textit{Last} accuracy are much lower compared with \textit{Upper-bound} due to the lack of training data for learned tasks during the continual learning process. All existing methods achieve significant improvement compared with \textit{Fine-tune} especially for \textit{ILIO}~\cite{ILIO}, which works more effectively when step size is very small as their final prediction is given by the combination of outputs for both the teacher model and current model. Note that \textit{ILIO} requires the teacher model for both training and inference phases which greatly increases the memory storage while other methods included ours only use teacher model during the training phase. However, as incremental step size increase, our method achieves best performance even for very large scale continual learning for $1,000$ classes in Food1K. We also show the accuracy evaluated after each incremental learning step with step size 5, 10, 20 and 100 in Figure~\ref{fig:sota}. Our method outperforms state-of-the-art for all learning steps with smallest performance gap compared with \textit{upper-bound}. Note that we did not provide the figures for step size 1 and 2 as they contain too many learning steps (100 and 50 respectively), which is difficult for visualization.

\begin{figure*}[t]
\begin{center}
  \includegraphics[width=1.\linewidth]{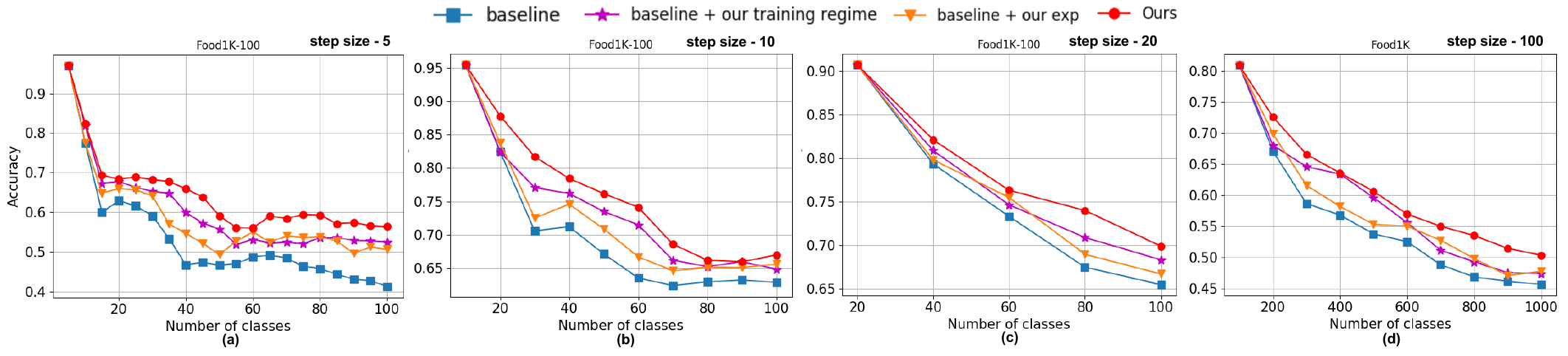}
  \vspace{-.6cm}
  \caption{\textbf{Ablation study} with step size (a) 5 (b) 10 (c) 20 on Food1K-100 and (d) step size 100 on Food-1K. (Best viewed in color)}
  \vspace{-0.3cm}
  \label{fig:ablation}
\end{center}
\end{figure*}

\begin{figure*}[t!]
\begin{center}
  \includegraphics[width=1.\linewidth]{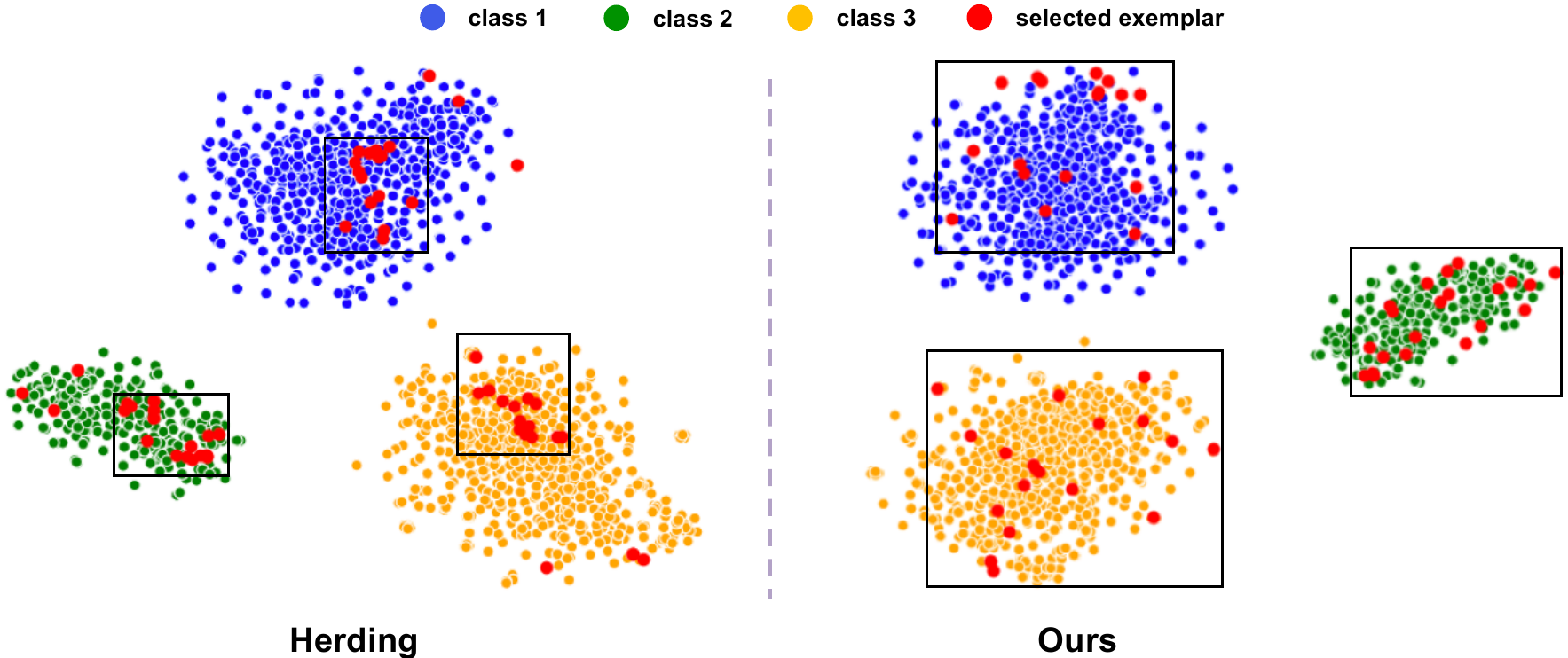}
  \vspace{-.5cm}
  \caption{\textbf{A t-SNE~\cite{t-sne} visualization} by comparing herding~\cite{HERDING} with our proposed exemplar selection algorithm. We randomly select three classes from Food1K corresponds to three different colors and the red dots represent the selected exemplars. The black box indicates the area where most exemplars are located for each class. (Best viewed in color)}
  \vspace{-0.3cm}
  \label{fig:tsne}
\end{center}
\end{figure*}

\begin{figure*}[t]
\begin{center}
  \includegraphics[width=1.\linewidth]{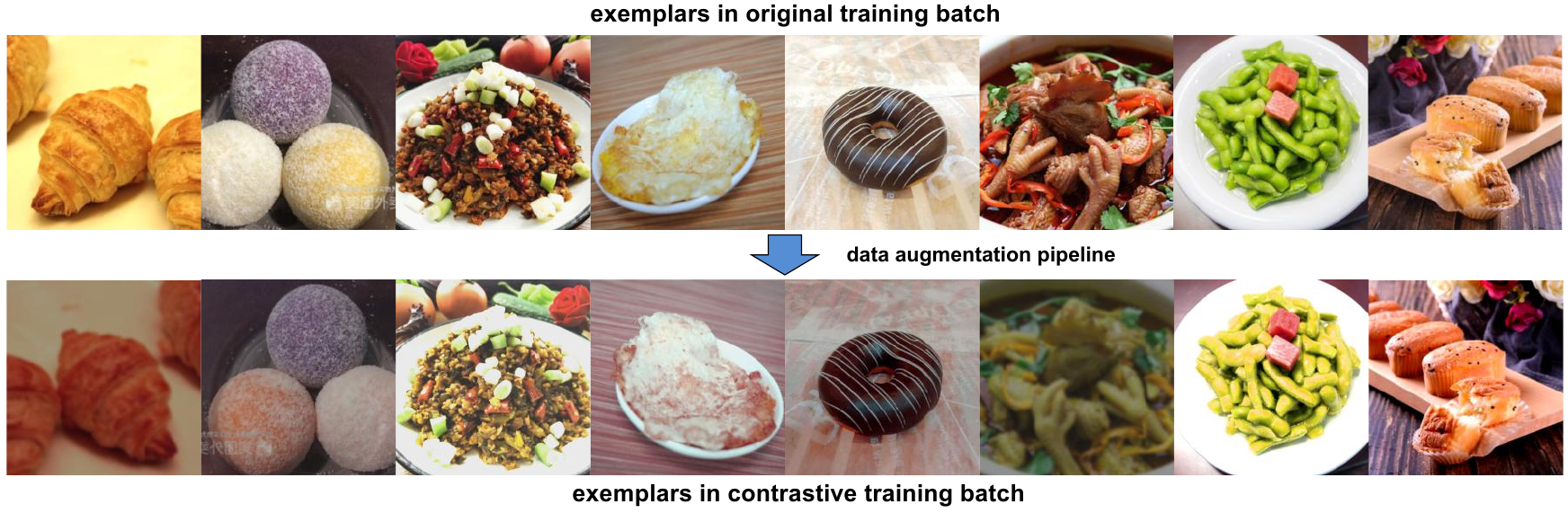}
  \vspace{-.2cm}
  \caption{\textbf{Visualization of contrastive training batch} generated by our proposed data augmentation pipeline including \textit{random flip}, \textit{random color distortions} and \textit{random Gaussian blur}. (Best viewed in color)}
  \vspace{-0.3cm}
  \label{fig:contrastive_batch}
\end{center}
\end{figure*}

\subsection{Ablation Study}
\label{sec: ablation study}
In this part, we conduct ablation studies to analyze the effectiveness of (1) \textbf{component-1:} our proposed exemplar selection algorithm that selects representative data from clusters generated based on visual similarity and (2) \textbf{component-2:} our online training regime using balanced training data for new and old class, and contrastive training batch for knowledge distillation. Specifically, we consider the following methods for comparisons:
\begin{itemize}
\itemsep0em 
  \item \textbf{baseline}: removing both component-1 and component-2 from our method, \textit{i.e.}, use herding~\cite{HERDING} for exemplar selection instead and pair new class data in training batch with the random number of exemplars
  \item \textbf{baseline + our exp}: baseline + component-1
  \item \textbf{baseline + our training regime}: baseline + component-2
  \item \textbf{Ours}: baseline + component-1 + component-2
\end{itemize}

Figure~\ref{fig:ablation} shows the results for each incremental step with step size 5, 10, 20 and 100. Compared with \textit{baseline}, we observe performance improvement by incorporating each component of proposed method. The best performance is obtained when combining both components. In addition, we notice that our training regime using balanced training batch performs more effectively than our exemplar selection since severe class-imbalanced problem exists in this Food1K dataset, where the number of training data ranges from $[91, 1199]$ per food class.

\begin{table}[t]
    \centering
    \scalebox{1.}{
    \begin{tabular}{|c|ccc|}
        \hline
            Method & $q=10$ & $q=50$ & $q=100$\\

        \hline
        \hline
        baseline  & 0.486 & 0.629 & 0.697 \\
        baseline + our exp & 0.527 & 0.651 & 0.706\\
        \hline
    \end{tabular}
    }
    \caption{\textbf{Average accuracy on Food1K-100 with step size 5 by varying exemplar size}. Best results marked in bold. }
    \label{tab:expsize}
\end{table}

\vspace{-0.3cm}
\subsubsection{Influence of Exemplar Size}
\label{sec:expsize}
For experiments in Section~\ref{sec:compare with SOTA}, we follow the protocol~\cite{ICARL} to use $20$ exemplars per class. In this part, we vary the number of exemplar stored for each class $q \in \{10, 50, 100\}$ and compare \textit{baseline + our exp} using our proposed exemplar selection algorithm with \textit{baseline} using Herding selection~\cite{HERDING}. We use Food1K-100 with step size $5$ and the average accuracy are shown in Table~\ref{tab:expsize}. In general, the performance becomes better for both methods when more exemplars are used. However, the memory storage capacity is one of the most important factors for continual learning especially in online scenario and we observe that our proposed approach is more efficient which outperforms \textit{baseline} for a larger margin when using less exemplars. 

\subsubsection{Visualization of Selected Exemplars}
\label{sec:vis of exp}
A t-SNE~\cite{t-sne} visualization comparing herding~\cite{HERDING} and our proposed exemplar selection method is shown in Figure~\ref{fig:tsne} where we randomly select three food classes from Food1K as denoted by blue, green and orange dots, respectively and red dots refer to the selected exemplars. As shown in the left half of the figure, most exemplars selected by herding are concentrated in a small area for each class as indicated by the black box. Therefore, the model gradually forgets the knowledge outside the black box during the continual learning process, leading to catastrophic forgetting. Our method addressed this problem by performing clustering at first based on visual similarity and then select exemplars from all generated clusters to better represent the intra-class diversity for each food class as illustrated in Section~\ref{method:exp selection}. In the right half of this figure, we find that the exemplars selected by our method covers a wider region for each food class, which helps to produce higher quality classifiers to retain the learned knowledge due to better generalization ability of our selected exemplars as shown in Figure~\ref{fig:ablation} by comparing \textbf{baseline} with \textbf{baseline + our exp}. 

\subsubsection{Visualization of Contrastive Training Batch}
Figure~\ref{fig:contrastive_batch} shows the exemplars for learned food classes in original and contrastive training batch using our proposed data augmentation pipeline including \textit{random flip}, \textit{random color distortions} and \textit{random Gaussian blur}. By comparing results of \textbf{baseline} with \textbf{baseline + our training regime} as shown in Figure~\ref{fig:ablation}, we observe that using augmented data is more effective to help retain the already learned knowledge to achieve better performance. One explanation is that each exemplar stored in the exemplar set can be selected for more than once to pair with new class data during the online training phase, so the data augmentation step helps to improve the classifier's generalization ability to obtain higher accuracy on learned classes. In addition, the knowledge distillation term also becomes more efficient to maintain the performance for old classes by using balanced training batch for old and new class data and transferring the learned knowledge from teacher model using original training batch to the current model using contrastive training batch as formulated in Equation~\ref{eq:kd1}.

\section{Conclusion}
\label{conclusion}
In summary, we studied online continual learning for food image classification in this work and proposed a novel exemplar selection algorithm that selected representative data from each cluster generated based on visual similarity to alleviate the high intra-class variation problem of food images. In addition, an effective online learning regime was introduced using balanced training batch for old and new class and we proposed to apply knowledge distillation using contrastive training batch to help retain the learned knowledge. Our method achieved promising results on a challenging food dataset, Food1K, with significant performance improvement compared with existing state-of-the-art especially when the number of new food classes added for each incremental step increased, showing great potential for large scale continual learning of food image classification in real life. 

For future work, although achieving promising results, our method still require storing part of original learned data as exemplars for replay during continual learning, which may not be feasible in many real scenarios due to the privacy issue or memory constraint. One possible solution is to use class prototype as recently introduced in~\cite{zhu2021prototype}.



{\small
\bibliographystyle{ieee_fullname}
\bibliography{egbib}

\begin{thebibliography}{10}\itemsep=-1pt

\bibitem{abati2020conditional_taskaware}
Davide Abati, Jakub Tomczak, Tijmen Blankevoort, Simone Calderara, Rita
  Cucchiara, and Babak~Ehteshami Bejnordi.
\newblock Conditional channel gated networks for task-aware continual learning.
\newblock {\em Proceedings of the IEEE/CVF Conference on Computer Vision and
  Pattern Recognition}, pages 3931--3940, 2020.

\bibitem{dualmemory}
Eden Belouadah and Adrian Popescu.
\newblock {Il2m}: Class incremental learning with dual memory.
\newblock {\em Proceedings of the IEEE International Conference on Computer
  Vision}, pages 583--592, 2019.

\bibitem{boushey2017new}
CJ Boushey, M Spoden, FM Zhu, EJ Delp, and DA Kerr.
\newblock New mobile methods for dietary assessment: review of image-assisted
  and image-based dietary assessment methods.
\newblock {\em Proceedings of the Nutrition Society}, 76(3):283--294, 2017.

\bibitem{EEIL}
Francisco~M. Castro, Manuel~J. Marin-Jimenez, Nicolas Guil, Cordelia Schmid,
  and Karteek Alahari.
\newblock End-to-end incremental learning.
\newblock {\em Proceedings of the European Conference on Computer Vision},
  September 2018.

\bibitem{ER_1}
Arslan Chaudhry, Marcus Rohrbach, Mohamed Elhoseiny, Thalaiyasingam Ajanthan,
  Puneet~K Dokania, Philip~HS Torr, and Marc'Aurelio Ranzato.
\newblock On tiny episodic memories in continual learning.
\newblock {\em arXiv preprint arXiv:1902.10486}, 2019.

\bibitem{douze2017evaluation}
Matthijs Douze, Herv{\'e} J{\'e}gou, and Jeff Johnson.
\newblock An evaluation of large-scale methods for image instance and class
  discovery.
\newblock {\em Proceedings of the on Thematic Workshops of ACM Multimedia},
  pages 1--9, 2017.

\bibitem{ER_2}
Tyler~L Hayes, Nathan~D Cahill, and Christopher Kanan.
\newblock Memory efficient experience replay for streaming learning.
\newblock {\em Proceedings of the International Conference on Robotics and
  Automation}, pages 9769--9776, 2019.

\bibitem{he2021end}
Jiangpeng He, Runyu Mao, Zeman Shao, Janine~L. Wright, Deborah~A. Kerr,
  Carol~J. Boushey, and Fengqing Zhu.
\newblock An end-to-end food image analysis system.
\newblock {\em Electronic Imaging}, 2021(8):285--1--285--7, 2021.

\bibitem{ILIO}
Jiangpeng He, Runyu Mao, Zeman Shao, and Fengqing Zhu.
\newblock Incremental learning in online scenario.
\newblock {\em Proceedings of the IEEE Conference on Computer Vision and
  Pattern Recognition}, pages 13926--13935, 2020.

\bibitem{he2020multitask}
Jiangpeng He, Zeman Shao, Janine Wright, Deborah Kerr, Carol Boushey, and
  Fengqing Zhu.
\newblock Multi-task image-based dietary assessment for food recognition and
  portion size estimation.
\newblock {\em arXiv preprint arXiv:2004.13188}, 2020.

\bibitem{he2021unsupervised}
Jiangpeng He and Fengqing Zhu.
\newblock Unsupervised continual learning via pseudo labels.
\newblock {\em arXiv preprint arXiv:2104.07164}, 2021.

\bibitem{RESNET}
Kaiming He, Xiangyu Zhang, Shaoqing Ren, and Jian Sun.
\newblock Deep residual learning for image recognition.
\newblock {\em Proceedings of the IEEE Conference on Computer Vision and
  Pattern Recognition}, pages 770--778, 2016.

\bibitem{KD}
Geoffrey Hinton, Oriol Vinyals, and Jeffrey Dean.
\newblock Distilling the knowledge in a neural network.
\newblock {\em Proceedings of the NIPS Deep Learning and Representation
  Learning Workshop}, 2015.

\bibitem{hand_crafted2}
H. {Hoashi}, T. {Joutou}, and K. {Yanai}.
\newblock Image recognition of 85 food categories by feature fusion.
\newblock {\em Proceedings of 2010 IEEE International Symposium on Multimedia},
  pages 296--301, Dec 2010.

\bibitem{rebalancing}
Saihui Hou, Xinyu Pan, Chen~Change Loy, Zilei Wang, and Dahua Lin.
\newblock Learning a unified classifier incrementally via rebalancing.
\newblock {\em Proceedings of the IEEE Conference on Computer Vision and
  Pattern Recognition}, pages 831--839, 2019.

\bibitem{hsu2018re}
Yen-Chang Hsu, Yen-Cheng Liu, Anita Ramasamy, and Zsolt Kira.
\newblock Re-evaluating continual learning scenarios: A categorization and case
  for strong baselines.
\newblock {\em arXiv preprint arXiv:1810.12488}, 2018.

\bibitem{Kagaya}
Hokuto Kagaya, Kiyoharu Aizawa, and Makoto Ogawa.
\newblock Food detection and recognition using convolutional neural network.
\newblock {\em Proceedings of the 22nd ACM International Conference on
  Multimedia}, pages 1085--1088, 2014.
\newblock {Orlando, Florida, USA}.

\bibitem{CIFAR}
Alex Krizhevsky, Geoffrey Hinton, et~al.
\newblock Learning multiple layers of features from tiny images.
\newblock 2009.

\bibitem{MNIST}
Yann LeCun, Corinna Cortes, and CJ Burges.
\newblock Mnist handwritten digit database.
\newblock {\em ATT Labs [Online]. Available: http://yann.lecun.com/exdb/mnist},
  2, 2010.

\bibitem{inthewild}
Kibok Lee, Kimin Lee, Jinwoo Shin, and Honglak Lee.
\newblock Overcoming catastrophic forgetting with unlabeled data in the wild.
\newblock {\em Proceedings of the IEEE International Conference on Computer
  Vision}, pages 312--321, 2019.

\bibitem{LWF}
Zhizhong Li and Derek Hoiem.
\newblock Learning without forgetting.
\newblock {\em IEEE Transactions on Pattern Analysis and Machine Intelligence},
  40(12):2935--2947, 2017.

\bibitem{PIC}
Frank Lin and William~W Cohen.
\newblock Power iteration clustering.
\newblock {\em Proceedings of International Conference on Machine Learning},
  2010.

\bibitem{lin2021most}
Luotao Lin, Fengqing Zhu, Edward Delp, and Heather Eicher-Miller.
\newblock The most frequently consumed and the largest energy contributing
  foods of us insulin takers using nhanes 2009--2016.
\newblock {\em Current Developments in Nutrition}, 5:426--426, 2021.

\bibitem{K-MEANS}
Stuart Lloyd.
\newblock Least squares quantization in pcm.
\newblock {\em IEEE transactions on information theory}, 28(2):129--137, 1982.

\bibitem{SIFT}
David~G Lowe.
\newblock Object recognition from local scale-invariant features.
\newblock {\em Proceedings of the seventh IEEE international conference on
  computer vision}, 2:1150--1157, 1999.

\bibitem{SIT}
Davide Maltoni and Vincenzo Lomonaco.
\newblock Continuous learning in single-incremental-task scenarios.
\newblock {\em Neural Networks}, 116:56--73, 2019.

\bibitem{mao2020visual}
Runyu Mao, Jiangpeng He, Zeman Shao, Sri~Kalyan Yarlagadda, and Fengqing Zhu.
\newblock Visual aware hierarchy based food recognition.
\newblock {\em arXiv preprint arXiv:2012.03368}, 2020.

\bibitem{wide}
N. {Martinel}, G.~L. {Foresti}, and C. {Micheloni}.
\newblock Wide-slice residual networks for food recognition.
\newblock {\em Proceedings of IEEE Winter Conference on Applications of
  Computer Vision}, pages 567--576, March 2018.

\bibitem{CF}
Michael McCloskey and Neal~J Cohen.
\newblock Catastrophic interference in connectionist networks: The sequential
  learning problem.
\newblock In {\em Psychology of Learning and Motivation}, volume~24, pages
  109--165. Elsevier, 1989.

\bibitem{NCM_classifier}
Thomas Mensink, Jakob Verbeek, Florent Perronnin, and Gabriela Csurka.
\newblock Distance-based image classification: Generalizing to new classes at
  near-zero cost.
\newblock {\em IEEE transactions on pattern analysis and machine intelligence},
  35(11):2624--2637, 2013.

\bibitem{min2020isia-500}
Weiqing Min, Linhu Liu, Zhiling Wang, Zhengdong Luo, Xiaoming Wei, Xiaolin Wei,
  and Shuqiang Jiang.
\newblock Isia food-500: A dataset for large-scale food recognition via stacked
  global-local attention network.
\newblock {\em Proceedings of the 28th ACM International Conference on
  Multimedia}, pages 393--401, 2020.

\bibitem{Food2K}
Weiqing Min, Zhiling Wang, Yuxin Liu, Mengjiang Luo, Liping Kang, Xiaoming Wei,
  Xiaolin Wei, and Shuqiang Jiang.
\newblock Large scale visual food recognition.
\newblock {\em CoRR}, abs/2103.16107, 2021.

\bibitem{pytorch}
Adam Paszke, Sam Gross, Soumith Chintala, Gregory Chanan, Edward Yang, Zachary
  DeVito, Zeming Lin, Alban Desmaison, Luca Antiga, and Adam Lerer.
\newblock Automatic differentiation in {PyTorch}.
\newblock {\em Proceedings of the Advances Neural Information Processing
  Systems Workshop}, 2017.

\bibitem{prabhu2020gdumb_online}
Ameya Prabhu, Philip~HS Torr, and Puneet~K Dokania.
\newblock Gdumb: A simple approach that questions our progress in continual
  learning.
\newblock {\em Proceedings of the European Conference on Computer Vision},
  pages 524--540, 2020.

\bibitem{ICARL}
Sylvestre-Alvise Rebuffi, Alexander Kolesnikov, Georg Sperl, and Christoph~H.
  Lampert.
\newblock {iCaRL}: Incremental classifier and representation learning.
\newblock {\em Proceedings of the IEEE Conference on Computer Vision and
  Pattern Recognition}, July 2017.

\bibitem{shao2021towards}
Zeman Shao, Shaobo Fang, Runyu Mao, Jiangpeng He, Janine Wright, Deborah Kerr,
  Carol~Jo Boushey, and Fengqing Zhu.
\newblock Towards learning food portion from monocular images with cross-domain
  feature adaptation.
\newblock {\em arXiv preprint arXiv:2103.07562}, 2021.

\bibitem{Tanno2016}
Ryosuke Tanno, Koichi Okamoto, and Keiji Yanai.
\newblock Deepfoodcam: A dcnn-based real-time mobile food recognition system.
\newblock {\em Proceedings of the 2Nd International Workshop on Multimedia
  Assisted Dietary Management}, pages 89--89, 2016.

\bibitem{t-sne}
Laurens Van~der Maaten and Geoffrey Hinton.
\newblock Visualizing data using t-sne.
\newblock {\em Journal of machine learning research}, 9(11), 2008.

\bibitem{vitter1985random}
Jeffrey~S Vitter.
\newblock Random sampling with a reservoir.
\newblock {\em ACM Transactions on Mathematical Software}, 11(1):37--57, 1985.

\bibitem{HERDING}
Max Welling.
\newblock Herding dynamical weights to learn.
\newblock {\em Proceedings of the International Conference on Machine
  Learning}, pages 1121--1128, 2009.

\bibitem{wu2016learning}
Hui Wu, Michele Merler, Rosario Uceda-Sosa, and John~R Smith.
\newblock Learning to make better mistakes: Semantics-aware visual food
  recognition.
\newblock {\em Proceedings of the 24th ACM international conference on
  Multimedia}, pages 172--176, 2016.

\bibitem{BIC}
Yue Wu, Yinpeng Chen, Lijuan Wang, Yuancheng Ye, Zicheng Liu, Yandong Guo, and
  Yun Fu.
\newblock Large scale incremental learning.
\newblock {\em Proceedings of the IEEE Conference on Computer Vision and
  Pattern Recognition}, June 2019.

\bibitem{mainatining}
Bowen Zhao, Xi Xiao, Guojun Gan, Bin Zhang, and Shu-Tao Xia.
\newblock Maintaining discrimination and fairness in class incremental
  learning.
\newblock {\em Proceedings of the IEEE Conference on Computer Vision and
  Pattern Recognition}, pages 13208--13217, 2020.

\bibitem{zhu2021prototype}
Fei Zhu, Xu-Yao Zhang, Chuang Wang, Fei Yin, and Cheng-Lin Liu.
\newblock Prototype augmentation and self-supervision for incremental learning.
\newblock {\em Proceedings of the IEEE Conference on Computer Vision and
  Pattern Recognition}, pages 5871--5880, 2021.

\end{thebibliography}
}

\end{document}